\documentclass{article}
\usepackage{ijcai21}

\usepackage{times}
\usepackage{soul}
\usepackage{url}
\usepackage[hidelinks]{hyperref}
\usepackage[utf8]{inputenc}
\usepackage[small]{caption}
\usepackage{graphicx}
\usepackage{amsmath}
\usepackage{amsthm}
\usepackage{booktabs}
\usepackage{algorithm}

 \usepackage{multirow}
\usepackage{amsmath}
\usepackage{amssymb}
\usepackage{algpseudocode}
\usepackage{wrapfig}
\usepackage{mathrsfs}
\usepackage{subfigure}

\newcommand{\argmax}{\mathop{\mathrm{argmax}}}
\newcommand{\argmin}{\mathop{\mathrm{argmin}}}
\title{Graph-Free Knowledge Distillation for Graph Neural Networks\footnote{Code: https://github.com/Xiang-Deng-DL/GFKD} }

\author{
Xiang Deng
\and
Zhongfei Zhang
\affiliations
State University of New York at Binghamton\\
\emails
xdeng7@binghamton.edu,
zhongfei@cs.binghamton.edu
}

\begin{document}

\maketitle

\begin{abstract}
Knowledge distillation (KD) transfers knowledge from a teacher network to a student by enforcing the student to mimic the outputs of the pretrained teacher on training data.
However, data samples are not always accessible in many cases due to large data sizes, privacy, or confidentiality.
Many efforts have been made on addressing this problem for convolutional neural networks (CNNs) whose inputs lie in a grid domain within a continuous space such as images and videos, but largely overlook graph neural networks (GNNs) that handle non-grid data with different topology structures within a discrete space.
The inherent differences between their inputs make these CNN-based approaches not applicable to GNNs.
In this paper, we propose to our best knowledge the first dedicated approach to distilling knowledge from a GNN without graph data.
The proposed graph-free KD (GFKD) learns graph topology structures for knowledge transfer by modeling them with multivariate Bernoulli distribution.
We then introduce a gradient estimator to optimize this framework.
Essentially, the gradients w.r.t. graph structures are obtained by only using GNN forward-propagation without back-propagation, which means that GFKD is compatible with modern GNN libraries such as DGL and Geometric.
Moreover, we provide the strategies for handling different types of prior knowledge in the graph data or the GNNs.
Extensive experiments demonstrate that GFKD achieves the state-of-the-art performance for distilling knowledge from GNNs without training data.
\end{abstract}

\section{Introduction}
Knowledge Distillation (KD)~\cite{hinton2015distilling} aims to transfer useful knowledge from a teacher network to a student.
The effectiveness of KD to boost the student performance has been demonstrated across a wide range of applications in artificial intelligence~\cite{romero2014fitnets}.
As the knowledge in a teacher concentrates on a narrow manifold instead of the full input space, KD has a strong assumption that either the training dataset or some representative samples are available.
The requirement for observable data highly
limits its applications on data-unavailable cases.
For example, a deep model (e.g., ResNet-152) that is pretrained on a large-scale dataset of billions of data samples is released online.
One may wish to distill the knowledge from this powerful model into a compact and fast model for the deployment on resource-limited devices, which requires the access to the training dataset.
However, the dataset is not publicly available as it is not only large but also difficult to store and transfer.
In reality, it is not rare that a corporation or a leading research group shares their pretrained models while they do not release the training data due to large sizes, privacy, confidentiality, or security in medical or industrial domains.
\par

\begin{figure}
\setlength{\belowcaptionskip}{-0.5cm}
\centering
     \includegraphics[width=0.46\textwidth]{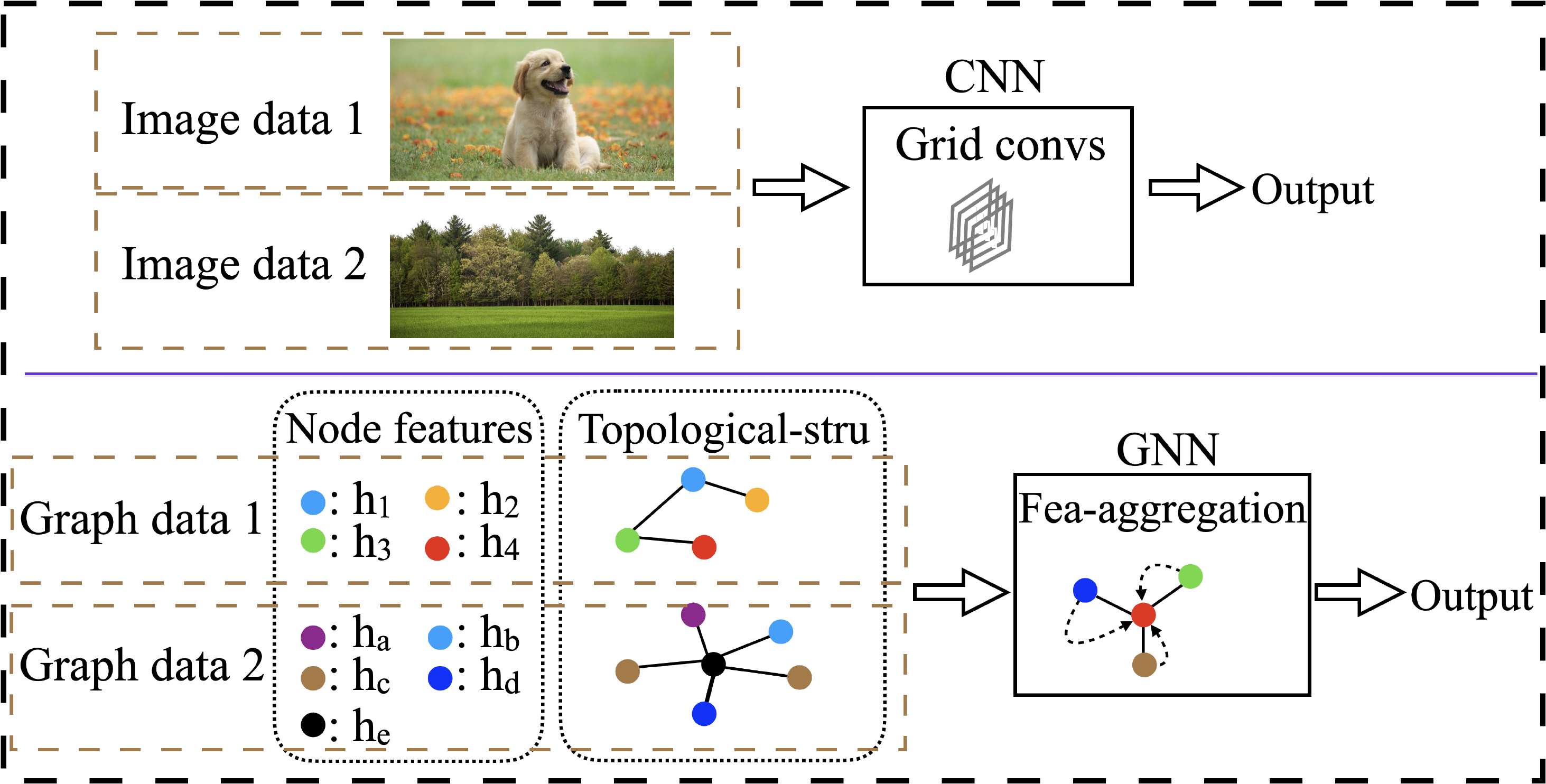}
     \caption{CNNs with grid data and GNNs with graph data}
     \label{cnngcn}
\end{figure}
A simple yet effective way to address the issue is to generate fake data for knowledge transfer by optimizing the inputs to the pretrained teacher to maximize the class-conditional probability~\cite{mordvintsev2015inceptionism}.
This method~\cite{yin2020dreaming} has shown its success on convolutional neural networks (CNNs) where the inputs are grid data within a continuous space such as images and videos as the gradients w.r.t. the inputs exist.
However, many real data such as proteins and chemical molecules lie in non-grid domains within a discrete space and thus call for graph neural networks (GNNs)~\cite{kipf2016semi} that explicitly deal with the topological structures of these graph data.
As shown in Figure \ref{cnngcn}, different from CNNs handling grid data such as images, GNNs deal with graph data that contain both node features and topological structures within a discrete space.
Current GNN models learn the node-level or graph-level representations by aggregating node features based on local topology structures.
The output of a GNN is not differentiable w.r.t. the topological structures of the input graphs, which makes the CNN-based approaches not applicable to GNNs.
\par

In this paper, we study how to distill knowledge from a pretrained GNN without observable graphs and develop to the best of our knowledge the first data-free knowledge distillation approach (i.e., GFKD) tailored for GNNs.
The workflow of GFKD is shown in Figure \ref{framework}.
GFKD first learns the fake graphs that the knowledge in the teacher GNN is more likely to concentrate on, and then uses these fake graphs to transfer knowledge to the student.
To achieve this goal, we propose a structure learning strategy by modeling the topology of a graph with a multivariate Bernoulli distribution and then introduce a gradient estimator to optimize it.\par

Our main contributions are summarized as follows:
\begin{itemize}
\item We introduce a novel framework, i.e., GFKD, for distilling knowledge from GNNs without observable graph data.
To the best of our knowledge, this is the first dedicated data-free KD approach tailored for GNNs.
We also provide the strategies (or regularizers) for handling different priors about the graph data, including how to deal with one-hot features and degree features.

\item
We develop a novel strategy for learning graph structures from a pretrained GNN by using multivariate Bernoulli distribution and introduce a gradient estimator to optimize it, which paves the way for extracting knowledge from a pretrained GNN without observable graphs.
Note that the current GNN libraries do not support computing gradients w.r.t. the input graph structures.
GFKD avoids this issue as the structure gradients in GFKD are obtained by only using GNN forward propagation without backward propagation.
GFKD is thus supported by these libraries.

\item We evaluate GFKD on six benchmark datasets in different domains with two different GNN architectures under different settings and demonstrate that GFKD achieves the best performance across different datasets.
\end{itemize}

\section{Related Work}

\subsection{Graph Neural Networks}
GNNs emerge as a hot topic in recent years for their potential in numerous applications.
Bruna et al.~\shortcite{bruna2013spectral} first generalize CNNs to signals defined on the non-grid domain.
Defferrard et al.~\shortcite{defferrard2016convolutional} further improve the idea by using Chebyshev polynomials. 
Kipf and Welling~\shortcite{kipf2016semi} propose to build GNNs by stacking multiple first-order Chebyshev polynomial filters.
Xu et al.~\shortcite{xu2018powerful}, on the other hand, propose a simple neural architecture, graph isomorphism networks (GINs), which generalizes the Weisfeiler-Lehman (WL) test and hence achieves a powerful discriminative ability.
The other GNN architectures including but not limited to~\cite{fey2018splinecnn,li2015gated,hamilton2017inductive,velivckovic2017graph} deal with the features and the topological structures using different strategies.

\subsection{Knowledge Distillation}
Knowledge distillation aims at transferring knowledge from a teacher model to a student.
Hinton et al.~\shortcite{hinton2015distilling} propose KD that penalizes the softened logit differences between a teacher and a student.
FitNet~\cite{romero2014fitnets} and AT~\cite{zagoruyko2016paying} further use the feature alignment to assist knowledge transfer.
Yang et al.~\shortcite{yang2020distilling} develop a local structure preserving module for distilling knowledge from GNNs.
Other distillation approaches~\cite{tian2019contrastive} use different criteria to align feature representations.
All these approaches require the training dataset for knowledge transfer, which cannot handle the case where the training data are not available.
\par
To address this issue, many efforts have been made on distilling knowledge without training images from a CNN through generating fake images.
Lopes et al.~\shortcite{lopes2017data} make use of meta data instead of real images to distill knowledge.
Yoo et al.~\shortcite{yoo2019knowledge} use a generator and a decoder to learn fake images for knowledge transfer.
Micaelli and Storkey~\shortcite{micaelli2019zero} and Chen et al~\shortcite{chen2019data} use generative adversarial networks (GANs)~\cite{goodfellow2014generative} to generate fake images.
Nayak et al.~\shortcite{nayak2019zero} propose to generate images by modelling the softmax space.
DeepInversion~\cite{yin2020dreaming} generates fake images by inversing a CNN and using the statistics in batch normalization~\cite{ioffe2015batch}.\par

All these data-free KD approaches are designed for CNNs with images as inputs, and highly overlook GNNs dealing with non-grid data within a discrete space.
Directly applying these approaches to GNNs is not applicable as graph data contain both features and topological structures.
Thus, it is necessary and appealing to develop a data-free KD approach tailored for GNNs.

\begin{figure}
\setlength{\belowcaptionskip}{-0.3cm}
\centering
     \includegraphics[width=0.47\textwidth]{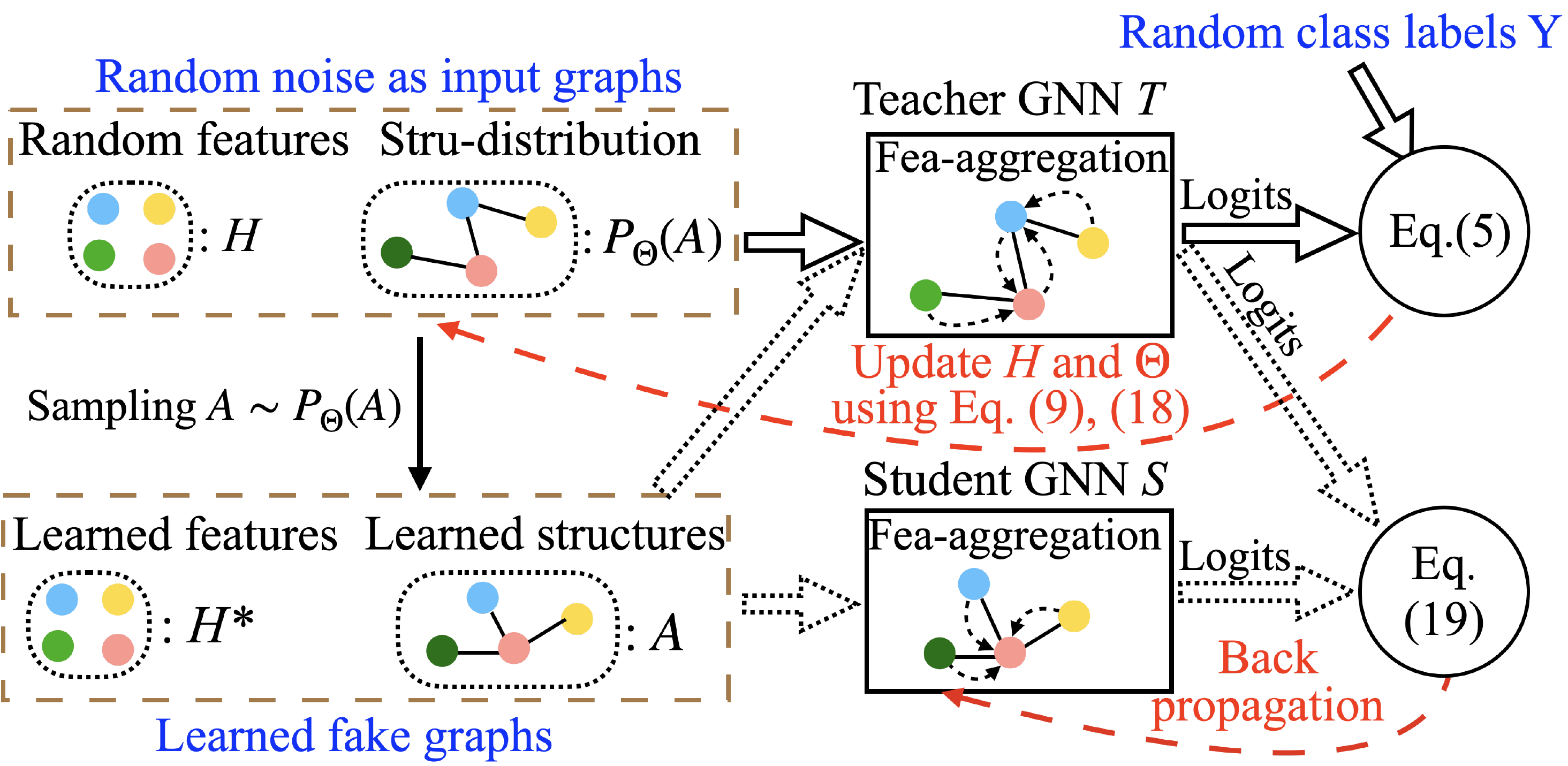}
     \caption{Framework of GFKD}
     \label{framework}
\end{figure}

\section{Framework}
In this section, we first provide a brief overview on GNNs.
We then present GFKD and the strategies for dealing with different types of prior knowledge about the graph data.
At the end, we introduce the optimization solution to GFKD.


\subsection{Graph Neural Networks}
Different from CNNs handling grid data, GNNs can take non-gird data as inputs.
A non-grid data can be represented as a graph $G = \{V, E\}$ and a set of features $h$, where $V$ and $E$ denote the nodes and the edges, respectively.
Modern GNNs typically adopt a neighborhood aggregation strategy by using the features and graph structures to learn discriminative node or graph representations.  
Suppose that $h_i\in h$ is the feature of node $v_i$.
The operation in layer $k$ of a GNN is:
\begin{equation}
\label{gcn}
h_i^k = g\left(f\left(h_{i}^{k-1}\right), \{f(h_{j}^{k-1})| v_j\in \mathcal{N}_i\}\right)
\end{equation}
where $g(.)$ denotes the aggregation function; $f(.)$ is the feature transformation function; $\mathcal{N}_i$ denotes the neighbors of node $v_i$.
It is observed that the topological structure of a graph plays a vital role in learning representations.
To transfer the knowledge from a teacher GNN to a student, it is necessary to know which structures the knowledge concentrates on.

\subsection{Graph-Free Knowledge Distillation}

Suppose that a teacher GNN $T(.)$ with parameters $W$ is trained over dataset (X, Y) by minimizing the regular cross-entropy loss, where X and Y are the graph data and the labels, respectively:
\begin{equation}
\label{l}
\mathcal{L}_{CE} = \mathcal{C}(Y, T_W(H, A))
\end{equation}
where $\mathcal{C}(.)$ denotes the loss function such as cross-entropy or mean square error; $H$ denotes the node features of graphs X; $A$ represents the graph structure information of X which can be represented as adjacency matrices consisting of 0s and 1s.

From the Bayesian perspective, learning $W$ by minimizing (\ref{l}) can be considered as maximizing the logarithm of class-conditional probability $\rm \log$ $p(Y|H, A, W)$, i.e., $\argmin_W\mathcal{L}_{CE}$ = $\argmax_W$ $\rm \log$ $p(Y|H, A, W)$.
Thus, the knowledge is more likely to concentrate on the graphs that make the teacher output a high class-conditional probability.
When training data are not available but $W$ is known, one may optimize the inputs by maximizing the class-conditional probability (i.e., $\argmax_{H, A}$ $\rm log$ $p(Y|H, A, W)$ or $\argmin_{H, A}\mathcal{L}_{CE}$) to generate fake samples for knowledge transfer.
This inversion technique~\cite{yin2020dreaming} has shown its success on CNNs where the inputs are grid-data images within a continuous space without topological structures $A$.
Unfortunately, this is not applicable to GNNs as $\mathcal{L}_{CE}$ is not differentiable w.r.t. the graph structure $A$ (but differentiable w.r.t. features $H$) as seen from (\ref{gcn}).


\subsubsection{Learning Graph Topology with Stochastic Structures}
To address the above issue, we propose to model the topological structure of a graph with a multivariate Bernoulli distribution, thus obtaining a stochastic structure.
Specifically, suppose that $a \in A$ is the structure (i.e., adjacency matrix) of a graph.
Each element in $a$ follows a Bernoulli distribution:
\begin{equation}
\label{ber}
P_{\theta_{ij}}(a_{ij}) = a_{ij}\phi(\theta_{ij}) + (1-a_{ij})\phi(-\theta_{ij})
\end{equation}
where $\phi(\theta_{ij})$ is the sigmoid function $\frac{e^{\theta_{ij}}}{1+e^{\theta_{ij}}}$; $\theta_{ij}$ is a learnable parameter; $a_{ij} \in \{0, 1\}$ is the element in the $i$th row and the $j$th column in $a$ where $a_{ij}=1$ means that node $v_i$ is a neighbor of $v_j$ and vice versa.

For directed graphs, $n^2$ parameters are used to model the distribution of $a$:
\begin{equation}
 P_\theta(a) = \prod_{i=1}^n\prod_{j=1}^n P_{\theta_{ij}}(a_{ij})
\end{equation}
Note that for undirected graphs, the number of parameters is reduced to $\frac{n(n+1)}{2}$ as their adjacency matrices are symmetric.\par

For a batch of graphs, their structures $A$ are independent of each other and thus follow the joint distribution: $\prod_{i=1}^m P_{\theta_i}(a)$ where $m$ is the number of graphs and $\theta_i$ is the structure parameter for graph $i$ .
Instead of directly minimizing (\ref{l}) that is not differentiable w.r.t. $A$, we generate fake graph data by minimizing the following expectation\footnote{Assume that we have some priors about the magnitude of the number of graph data nodes.}:
\begin{equation}
\label{obj}
\mathcal{L}_{H, \Theta} = \mathop{\mathbb{E}}_{A\sim P_\Theta(A)} \left[\mathcal{C}\left(Y, T_W\left(H, A\right)\right) + \lambda*\mathcal{R}\right]
\end{equation}
where $P_\Theta(A)$ = $\prod_{i=1}^m P_{\theta_i}(a)$; $H$ represents the feature parameters; $Y$ is a set of randomly sampled labels; $\mathcal{R}$ denotes the regularizers for different priors that we have about the target task data; $\lambda$ is a balancing weight.
Thus, minimizing (\ref{obj}) w.r.t. $H$ and $\Theta$ can generate the graph data that maximize the GNN output class probability.
We omit $W$ in $T_{W}$ in the following as it is known for a pretrained teacher GNN.

\subsubsection{Regularizers for Prior Knowledge}
$\mathcal{R}$ in (\ref{obj}) deals with different types of prior knowledge about the graph data for the target task.
We provide the strategies for handling common priors.
\par
\paragraph{Priors in Graph Neural Networks:}
Similar to the case in CNNs, many GNNs also benefit from batch normalization (BN).
BN contains statistical information about the data as it accumulates the moving average of the means and the variances of the features during training.
Similar to the case in CNNs~\cite{yin2020dreaming}, it is reasonable to force the fake graph data to have similar feature means and variances to those in the GNNs accumulated from the real graph data:
\begin{equation}
\label{objbn}
\mathcal{R}_{bn}= (u_{A, \Theta}-u_{T})^2 + (v_{A, \Theta}-v_{T})^2
\end{equation}
where $u_{A, \Theta}$ and $v_{A, \Theta}$ represent the means and variances of the features of the generated graphs, respectively; $u_{T}$ and $v_{T}$ are the means and variances in the teacher GNN, respectively.
\par

\paragraph{Priors about Target-Task Graph Data:} Besides the priors embedded in the GNNs, one may have some prior knowledge about the target task data:
\paragraph{(1) One-hot features:} The graph data for many tasks have one-hot features, such as the classification task on MUTAG.
In this case, directly minimizing (\ref{obj}) cannot lead to one-hot features.
To address this issue, we first reparameterize $H$ in (\ref{obj}) with the softmax function $\sigma(\omega)$ where $\omega$ are learnable parameters and then minimize the entropy of $\sigma(\omega)$:
\begin{equation}
\label{one}
\mathcal{L}_{\omega, \Theta} = \mathop{\mathbb{E}}_{A\sim P_\Theta(A)} \left[\mathcal{C}\left(Y, T\left(\sigma(\omega), A\right)\right) + \lambda*Ent(\sigma(\omega))\right]
\end{equation}
where $Ent(.)$ denotes the entropy and $\sigma(\omega)$ can be seen as the instantiations of $H$.
\paragraph{(2) Degrees as features:} Some graph data use the degrees of the nodes as features.
In this case, the features can be derived from the adjacency matrix $A$.
It is thus not necessary to explicitly learn features $H$ and objective (\ref{obj}) is reduced to:
\begin{equation}
\label{objnf}
\mathcal{L}_{\Theta} = \mathop{\mathbb{E}}_{A\sim P_\Theta(A)} \left[\mathcal{C}\left(Y, T\left(A\right)\right) + \lambda*\mathcal{R}\right]
\end{equation}

We have discussed some common graph priors while there may be other priors for different graphs.
Fortunately, objective (\ref{obj}) is readily extended to different graph data.

\subsubsection{Optimization}
To minimize objective (\ref{obj}), we need to compute the gradients regarding $H$ and $\Theta$.
As the gradients of $\mathcal{C}(.)$ w.r.t. $H$ exist, we can easily estimate them by sampling from $P_\Theta(A)$:
\begin{equation}
\label{feature}
\begin{split}
\nabla_H \mathcal{L}_{H, \Theta}
= \nabla_H\mathop{\mathbb{E}}_{A\sim P_\Theta(A)} \left[\mathcal{C}\left(Y, T\left(H, A\right)\right) + \lambda*\mathcal{R}\right]\\
=\mathop{\mathbb{E}}_{A\sim P_\Theta(A)}\nabla_H \left[\mathcal{C}\left(Y, T\left(H, A\right)\right) + \lambda*\mathcal{R}\right]\\
=\frac{1}{N}\sum_{i=1}^N\nabla_H \left[\mathcal{C}\left(Y, T\left(H, A^i\right)\right) + \lambda*\mathcal{R}\right]
\end{split}
\end{equation}
where $A^i\sim P_\Theta(A)$ are independent and identically distributed (iid).

The difficulty lies in computing the gradients w.r.t. $\Theta$ that exist in the distribution $P_\Theta(A)$.
We introduce a gradient estimator \cite{yin2019arm,williams1992simple} to optimize $\Theta$, which is based on the reparametrization trick and REINFORCE.
\par
We omit $\lambda*\mathcal{R}$ in the objective (\ref{obj}) in the following for simplicity.
Bernoulli random variables $A$ in (\ref{obj}) can be reparameterized by two exponential random variables:
\begin{equation}
\label{re1}
\begin{split}
\mathcal{L}_{H, \Theta}=\mathop{\mathbb{E}}_{A\sim P_\Theta(A)} \mathcal{C}\left(Y, T\left(H, A\right)\right)\\
=\mathop{\mathbb{E}}_{B, M\sim \prod_{i=1}^{n^2} Ep(1)} \mathcal{C}\left(Y, T\left(H, \mathbf1_{[B\odot e^{-\frac{\Theta}{2}} < M\odot e^{\frac{\Theta}{2}}]}\right)\right)
\end{split}
\end{equation}
where $Ep(.)$ represents the Exponential distribution;
$\odot$ denotes the element-wise product;
$\mathbf1_{[.]}$ is the indicator function which equals to one if the argument is true and zero otherwise.\par

Note that as $B$ follows $\prod_{i=1}^{n^2} Ep(1)$, $Q$ = $B\odot e^{-\frac{\Theta}{2}}$ follows $\prod_{i=1}^{n^2} Ep(e^{\frac{\theta_i}{2}})$. 
Similarly, $S$ = $M\odot e^{\frac{\Theta}{2}}$ follows $\prod_{i=1}^{n^2}$ $Ep(e^{-\frac{\theta_i}{2}})$.
(\ref{re1}) can be further reparameterized as:
\begin{equation}
\label{ree}
\begin{split}
\mathcal{L}_{H, \Theta}=\mathop{\mathbb{E}}_{Q\sim \prod_{i=1}^{n^2} Ep(e^{\frac{\theta_i}{2}}), S\sim \prod_{i=1}^{n^2} Ep(e^{-\frac{\theta_i}{2}})} \\\left[\mathcal{C}\left(Y, T\left(H, \mathbf1_{[Q < S]}\right)\right)\right]
\end{split}
\end{equation}


Next we show how to obtain the gradients w.r.t $\Theta$.
Applying REINFORCE to (\ref{ree}) leads to:
\begin{equation}
\label{reee}
\begin{split}
\nabla_{\Theta} \mathcal{L}_{H, \Theta}=\mathop{\mathbb{E}}_{Q\sim \prod_{i=1}^{n^2} Ep(e^{\frac{\theta_i}{2}}), S\sim \prod_{i=1}^{n^2} Ep(e^{-\frac{\theta_i}{2}})}\\ \left[\mathcal{C}\left(Y, T\left(H, \mathbf1_{[Q < S]}\right)\right)\nabla_{\Theta}log\left(P_\Theta\left(Q\right)P_\Theta\left(S\right)\right)\right]\\
=\mathop{\mathbb{E}}_{Q\sim \prod_{i=1}^{n^2} Ep(e^{\frac{\theta_i}{2}}), S\sim \prod_{i=1}^{n^2} Ep(e^{-\frac{\theta_i}{2}})} \\
\left[\mathcal{C}\left(Y, T\left(H, \mathbf1_{[Q < S]}\right)\right)
\frac{1}{2}(S\odot e^{-\frac{\Theta}{2}}-Q\odot e^{\frac{\Theta}{2}})\right]
\end{split}
\end{equation}

As $Q$ = $B\odot e^{-\frac{\Theta}{2}}$ and $S$ = $M\odot e^{\frac{\Theta}{2}}$, (\ref{reee}) is equivalent to:
\begin{equation}
\label{reeee_2}
\begin{split}
\nabla_{\Theta} \mathcal{L}_{H, \Theta} = \mathop{\mathbb{E}}_{B, M\sim \prod_{i=1}^{n^2} Ep(1)}\\ \left[\mathcal{C}\left(Y, T\left(H, \mathbf1_{\left[B\odot e^{\frac{-\Theta}{2}} < M\odot e^{\frac{\Theta}{2}}\right]}\right)\right)\frac{1}{2}(M-B)\right]
\end{split}
\end{equation}

$B$ and $M$ can be further reparameterized as:
\begin{equation}
B=K \odot U, M=K \odot (1-U)
\end{equation}
where $U$ and $K$ follow $\prod_{i=1}^{n^2} Un(0,1)$ and $\prod_{i=1}^{n^2}$ $Gamma(2,1)$, respectively, and $Un()$ and $Gamma()$ denote the uniform distribution and the gamma distribution, respectively.
(\ref{reeee_2}) can be further reparameterized as:
\begin{equation}
\begin{split}
\label{ee18}
\nabla_{\Theta} \mathcal{L}_{H, \Theta}=\mathop{\mathbb{E}}_{U \sim \prod_{i=1}^{n^2} Un(0,1), K \sim \prod_{i=1}^{n^2} Gamma(2,1)}\\
\left[\mathcal{C}\left(Y, T\left(H, \mathbf1_{[U<\phi(\Theta)]}\right)\right) \frac{1}{2}(K-2K\odot U)\right]
\end{split}
\end{equation}
By applying Rao Blackwellization to (\ref{ee18}), $\nabla_{\Theta} \mathcal{L}_{H, \Theta}$ is equal to:
\begin{equation}
\begin{split}
\label{ee19}
\mathop{\mathbb{E}}_{U \sim \prod_{i=1}^{n^2} Un(0,1)} \left[\mathcal{C}\left(Y, T\left(H, \mathbf1_{[U<\phi(\Theta)]}\right)\right)(1-2U)\right]
\end{split}
\end{equation}
$U\sim \prod_{i=1}^{n^2}Un(0, 1)$ implies $(1-U)\sim \prod_{i=1}^{n^2}Un(0, 1)$.
Thus, $U$ in (\ref{ee19}) can be replaced with $1-U$ to obtain a new unbiased gradient estimator.
Taking the average of the new estimator and (\ref{ee19}) can further reduce the sampling variance \cite{yin2019arm}:
\begin{equation}
\begin{split}
\label{e201}
\nabla_{\Theta} \mathcal{L}_{H, \Theta}=\mathop{\mathbb{E}}_{U \sim \prod_{i=1}^{n^2} Un(0,1)} [(\mathcal{C}\left(Y, T\left(H, \mathbf1_{[U>\phi(-\Theta)]}\right)\right)\\-\mathcal{C}\left(Y, T\left(H, \mathbf1_{[U<\phi(\Theta)]}\right)\right))(U-0.5)]
\end{split}
\end{equation}


By simply sampling from $Un(0, 1)$, we can obtain the gradients w.r.t. $\Theta$
\begin{equation}
\begin{split}
\label{ee23}
\nabla_{\Theta} \mathcal{L}_{H, \Theta}=\frac{1}{N}\sum_{i=1}^N [(\mathcal{C}\left(Y, T\left(H, \mathbf1_{[U^i>\phi(-\Theta)]}\right)\right)\\-\mathcal{C}\left(Y, T\left(H, \mathbf1_{[U^i<\phi(\Theta)]}\right)\right))(U^i-0.5)]
\end{split}
\end{equation}
where $U^i\sim \prod_{i=1}^{n^2} Un(0,1)$ are iid.
More derivation details are given in the Appendix.
We set $N$ to 1 in this paper as the variance is small.
As shown in (\ref{ee23}), this unbiased gradient estimator allows us to obtain the gradients w.r.t. the structure parameters $\Theta$ with only GNN forward propagation and thus it is efficient and supported by the current GNN libraries. 

\begin{table*}[!t]
\centering
\resizebox{\textwidth}{!}{%
\begin{tabular}{lcccc|ccc|ccc}
\toprule
Datasets &                                                                                & \multicolumn{3}{c|}{MUTAG}         & \multicolumn{3}{c|}{PTC}       & \multicolumn{3}{c}{PROTEINS}    \\ \midrule
Teacher  &             & GCN-5-64  & GIN-5-64  & GCN-5-64   & GCN-5-64 & GIN-5-64 & GCN-5-64 & GCN-5-64 & GIN-5-64 & GCN-5-64  \\
Student  &                     & GCN-3-32  & GIN-3-32  & GIN-3-32   & GCN-3-32 & GIN-3-32 & GIN-3-32 & GCN-3-32 & GIN-3-32 & GIN-3-32  \\ \midrule
Teacher  &100\% training data & 89.5 & 92.9  & 89.5   & 63.5 & 61.5 & 63.5 & 76.9 & 76.0 & 76.9  \\
KD       &       100\% training data                              & 84.6$\pm$2.3  & 87.7$\pm$2.1  & 84.4$\pm$1.7   & 60.2$\pm$2.1 & 60.5$\pm$1.9 & 60.6$\pm$2.6 & 76.0$\pm$1.0 & 76.7$\pm$0.6 & 76.3$\pm$0.8  \\ \midrule
RandG    & 0 training data                                                                & 39.1$\pm$6.8 & 58.7$\pm$4.2 & 38.8$\pm$5.8  & 43.2$\pm$4.4 & 53.2$\pm$5.8 & 42.9$\pm$8.0 & 56.6$\pm$4.8 & 33.3$\pm$7.4 & 43.4$\pm$9.5 \\
DeepInvG   & 0 training data                                                                & 58.6$\pm$5.1  & 59.6$\pm$2.9  &35.4$\pm$2.2  & 52.9$\pm$7.8 & 45.7$\pm$4.1 & 43.9$\pm$9.2 & 65.4$\pm$2.3 & 52.7$\pm$5.7 & 47.4$\pm$9.9  \\
GFKD     & 0 training data                                                                & \textbf{70.8$\pm$2.1}  & \textbf{73.2$\pm$4.2} & \textbf{70.2$\pm$2.0} & \textbf{57.4$\pm$2.2} & \textbf{57.5$\pm$2.4} & \textbf{54.1$\pm$6.4} & \textbf{74.7$\pm$1.5} & \textbf{60.4$\pm$1.0} & \textbf{65.5$\pm$3.1}  \\ \bottomrule
\end{tabular}
}
\setlength{\abovecaptionskip}{0.1cm}
\caption{Test accuracies (\%) on MUTAG, PTC, and PROTEINS}
\label{bio}
\end{table*}

\begin{table*}[!t]
\centering
\resizebox{\textwidth}{!}{%
\begin{tabular}{lcccc|ccc|ccc}
\toprule
Datasets &                                                                                & \multicolumn{3}{c|}{IMDB-B}         & \multicolumn{3}{c|}{COLLAB}       & \multicolumn{3}{c}{REDDIT-B}    \\ \midrule
Teacher  &                                                                                & GCN-5-64  & GIN-5-64  & GCN-5-64   & GCN-5-64 & GIN-5-64 & GCN-5-64 & GCN-5-64 & GIN-5-64 & GCN-5-64  \\
Student  &                                                                                & GCN-3-32  & GIN-3-32  & GIN-3-32   & GCN-3-32 & GIN-3-32 & GIN-3-32 & GCN-3-32 & GIN-3-32 & GIN-3-32  \\ \midrule
Teacher  &100\% training data & 74.3  & 75.3  &74.3   & 81.7 & 82.3 & 81.7 & 92.8 &91.7 & 92.8  \\
KD       &       100\% training data                              & 76.4$\pm$0.8  & 75.4$\pm$0.9   & 75.6$\pm$1.3   &80.9$\pm$0.6 &81.9$\pm$0.5 &81.6$\pm$0.4 &85.7$\pm$0.6 &88.4$\pm$2.5 &87.2$\pm$3.4  \\ \midrule
RandG/DeepInvG    & 0 training data & 58.5$\pm$3.7 & 58.7$\pm$4.2 & 55.4$\pm$3.4  & 34.8$\pm$9.0 &28.4$\pm$7.3 & 27.2$\pm$6.3 & 50.1$\pm$1.0 & 49.9$\pm$0.8 & 48.9$\pm$2.1 \\
GFKD     & 0 training data                                                                &\textbf{69.2$\pm$1.1}  & \textbf{67.8$\pm$1.8} & \textbf{67.1$\pm$2.3} & \textbf{67.3$\pm$1.4} & \textbf{65.4$\pm$2.7} & \textbf{66.8$\pm$1.8} & \textbf{66.5$\pm$3.7} & \textbf{63.8$\pm$4.5} & \textbf{63.1$\pm$5.7}  \\ \bottomrule
\end{tabular}
}
\setlength{\abovecaptionskip}{0.1cm}
\caption{Test accuracies (\%) on IMDB-B, COLLAB, and REDDIT-B}
\label{socail}
\end{table*}

\begin{table}[!b] 
\vskip -0.01in
\centering
\begin{tabular}{cccc}
\toprule
        & \#Graphs & \#Classes & Avg\#Graph Size \\ \midrule
MUTAG   & 188      & 2         & 17.93           \\
PTC   & 344      & 2         & 14.29           \\
PROTEINS & 1,113      & 2         & 39.06      \\
  IMDB-B  & 1,000     & 2         & 19.77           \\
COLLAB & 5,000     & 3         & 74.49          \\
REDDIT-B &2,000 &2 &429.62\\
 \bottomrule
\end{tabular}
\setlength{\abovecaptionskip}{0.1cm}
\caption{Summary of datasets.}
\label{dataset}
\end{table}

\subsubsection{Knowledge Transfer with Generated Fake Graphs}
As shown in Figure \ref{framework}, we first update feature parameters $H$ and the stochastic structure parameters $\Theta$ by minimizing ($\ref{obj}$) with gradients (\ref{feature}) and (\ref{ee23}), respectively.
We then simply obtain fake graphs by using $H$ as the node features and sampling from $P_{\Theta}(A)$ to generate topological structures.
The teacher GNN outputs a high probability on these graphs and thus the knowledge is more likely to concentrate on these graphs.
We then transfer knowledge from the teacher to the student by using these fake graph data $x$ with the KL-divergence loss:
\begin{equation}
\label{kl}
 \tau^2 KL\left(\sigma\left(\frac{T\left(x\right)}{\tau}\right), \sigma\left(\frac{S\left(x\right)}{\tau}\right)\right) 
\end{equation}
where $\sigma$ is the softmax function; $\tau$ is a temperature to generate soft labels and $KL$ represents KL-divergence; $S$ is the student GNN.
To the end, we achieve knowledge distillation without any observable graphs.



\section{Experiments}
In this section, we report extensive experiments for evaluating GFKD.
Note that our goal is not to generate real graphs but to transfer as much knowledge as possible from a pretrained teacher GNN to a student without using any training data.

\subsection{Experimental Settings}
We adopt six graph classification benchmark datasets~\cite{xu2018powerful} including three bioinformatics graph datasets, i.e., MUTAG, PTC, and PROTEINS, and three social network graph datasets, i.e., IMDB-B, COLLAB, and REDDIT-B. 
The statistics of these datasets are summarized in Table \ref{dataset}.
On each dataset, 70\% data are used for pretraining the teachers and the remaining 30\% are used as the test data.
\par
We use two well known GNN architectures, i.e, GCN~\cite{kipf2016semi} and GIN~\cite{xu2018powerful}.
We distill knowledge in two different settings, i..e, the teacher and the student share the same architecture or use different architectures.
We use the form of (architecture-layer number-feature dimensions) to denote a GNN.
For example, GIN-5-64 denotes a GNN with 5 GIN layers and 64 feature dimensions.
\par

As there is no existing approach applicable to GNNs for distilling knowledge without observable graph data, we design two baselines for references:
\begin{itemize}
\item Random Graphs (RandG): RandG generates fake graphs by randomly drawing from a uniform distribution as node features and topological structures, and then use these graphs to transfer knowledge.
\item DeepInvG~\cite{yin2020dreaming}: As the original DeepInversion~\cite{yin2020dreaming} cannot learn the structures of graph data, here DeepInvG first randomly generates graph structures and then uses DeepInversion to learn node features with objective $C(Y, T(H, A))+\mathcal{R}_{bn}$.
\end{itemize}

For generating fake graphs, we do 2500 iterations.
The learning rates for structure and feature parameters are set to 1.0 (5.0 on PROTEINS, COLLAB, and REDDIT-B) and 0.01, respectively, and are divided by 10 every 1000 iterations.
For KD, all the GNNs are trained for 400 epochs with Adam and the learning rate is linearly decreased from 1.0 to 0.
$\tau$ is set to 2.
More details are given in the Appendix.

\begin{figure*}[!t]
     \begin{minipage}{0.47\textwidth}
      \centering
     \includegraphics[width=0.87\textwidth]{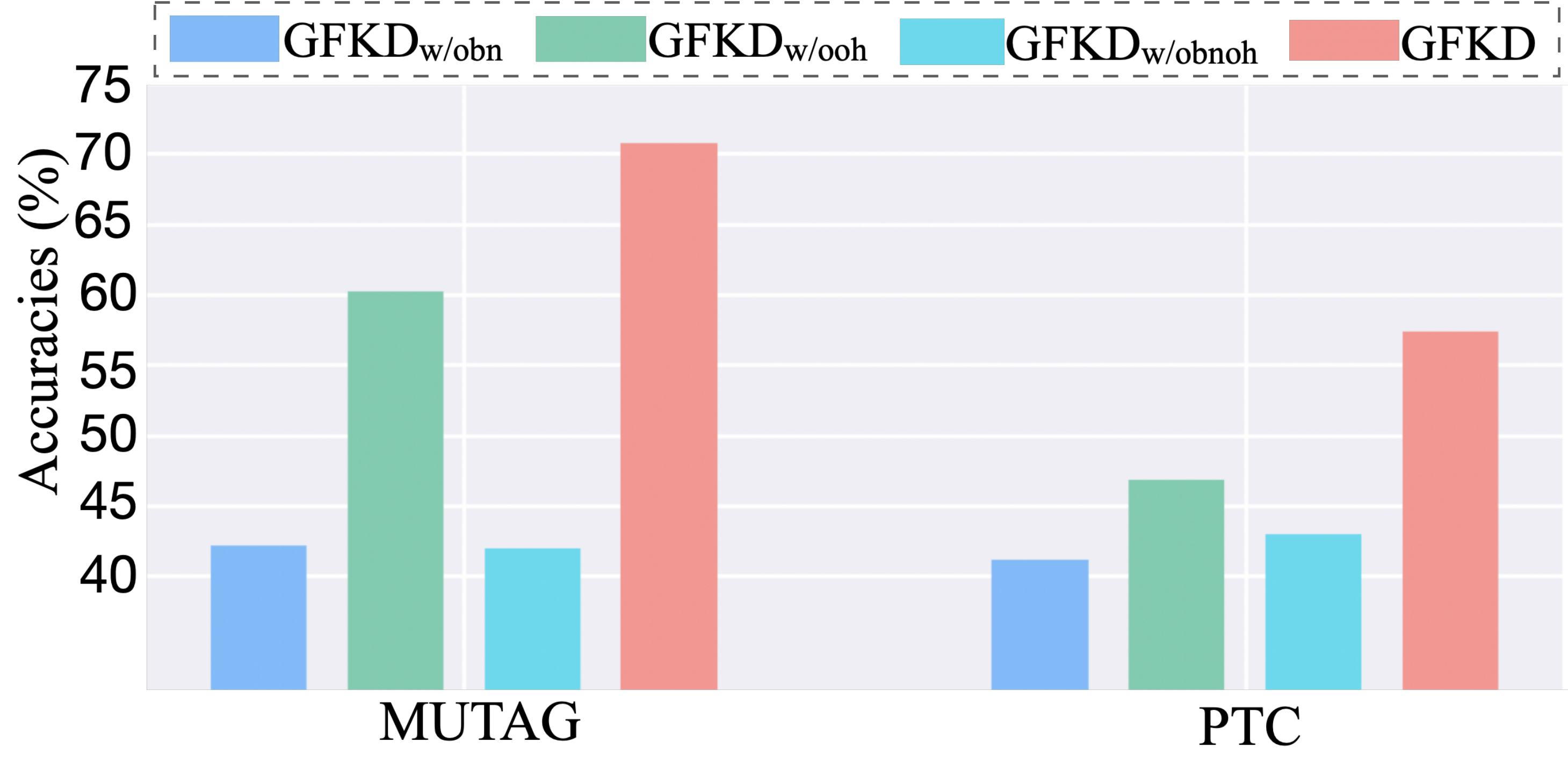}
     \vskip -0.1in
     \caption{Ablation studies regarding regularizers}
     \label{aba}
   \end{minipage}\hfill
   \begin{minipage}{0.47\textwidth}
\centering
     \includegraphics[width=0.97\textwidth]{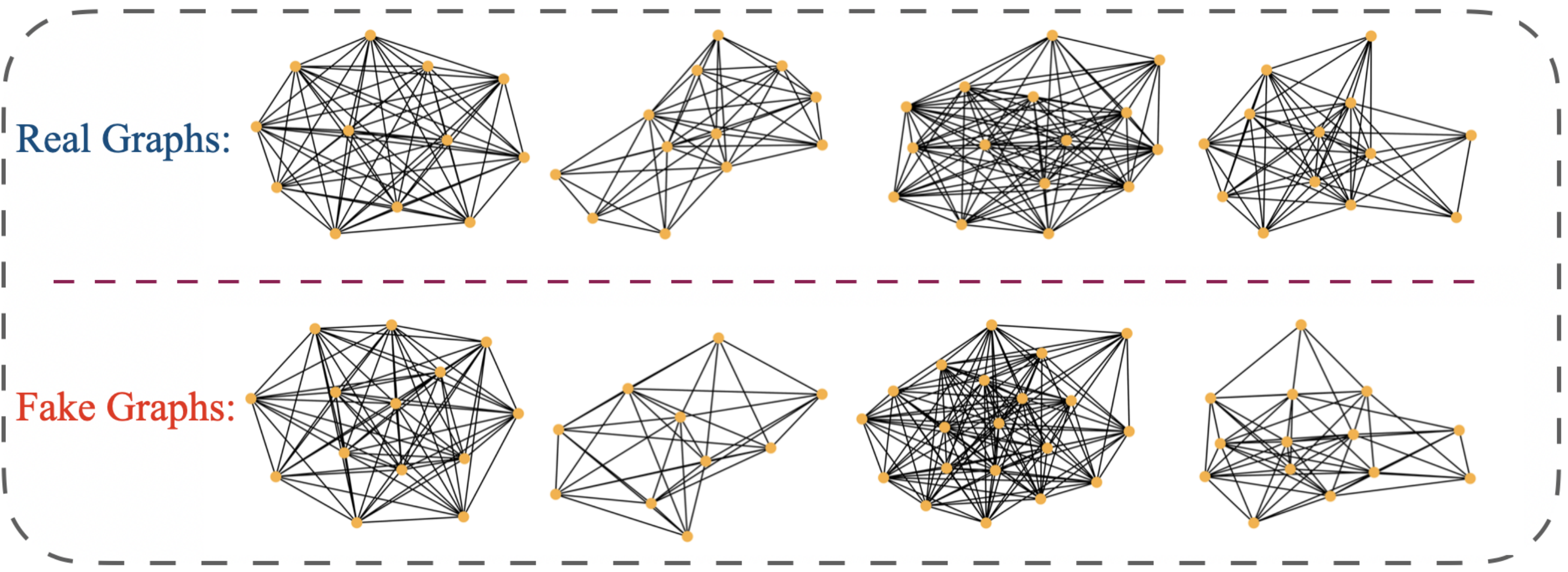}
     \caption{Graph visualization on IMDB-B. Note that there is no correspondence between the graphs in the two rows.}
     \label{vis}
   \end{minipage}\hfill
   \vskip -0.05in
\end{figure*}

\begin{figure*}
\begin{minipage}{0.24\textwidth}
\centering
\includegraphics[height=2.7 cm]{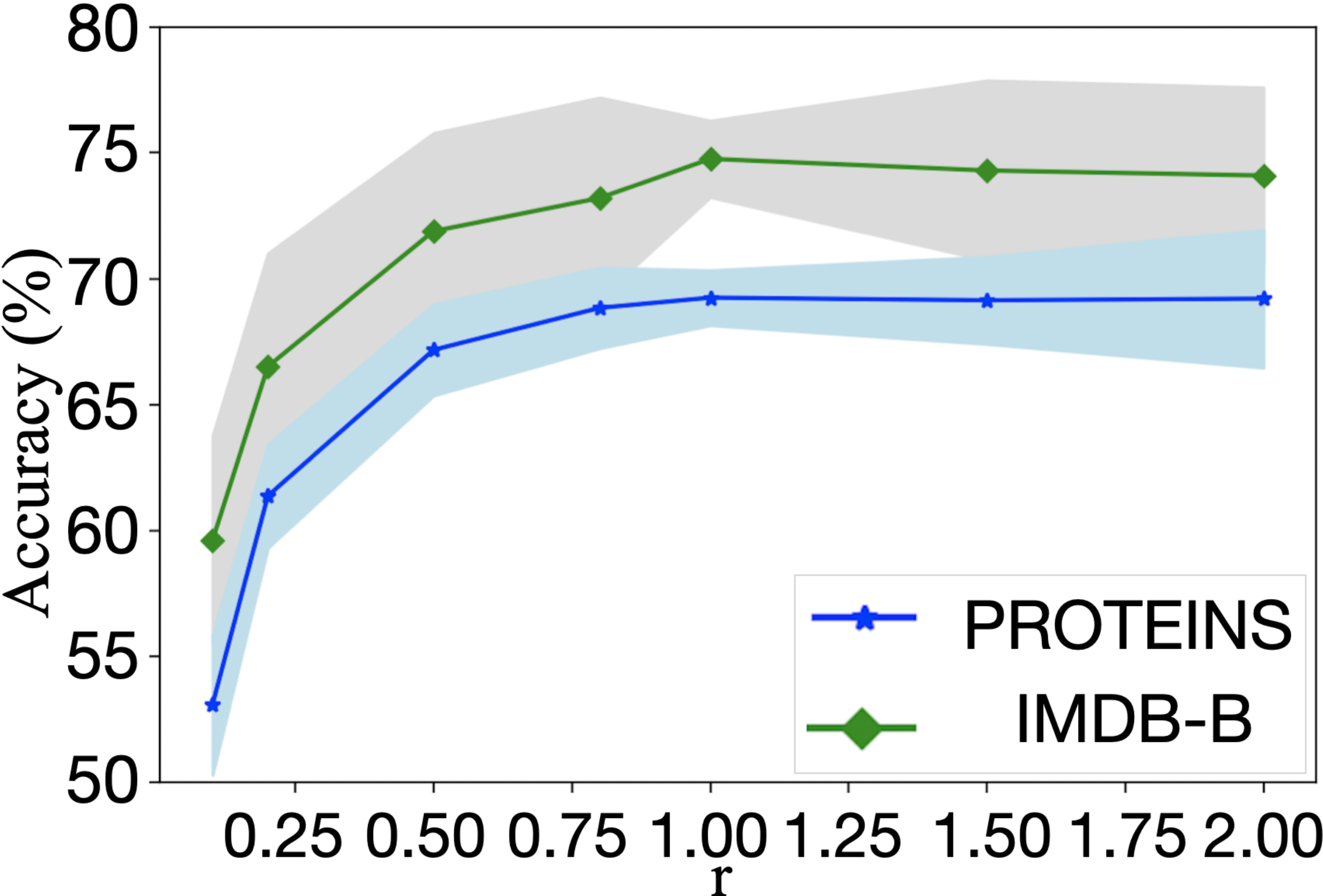}
\caption{Ablation studies regarding the number of fake graphs}
\label{number}
\end{minipage}
 \begin{minipage}{0.75\textwidth}
 \centering
 \subfigure[RandG]{\centering\label{fig:b}\includegraphics[height=2.6cm]{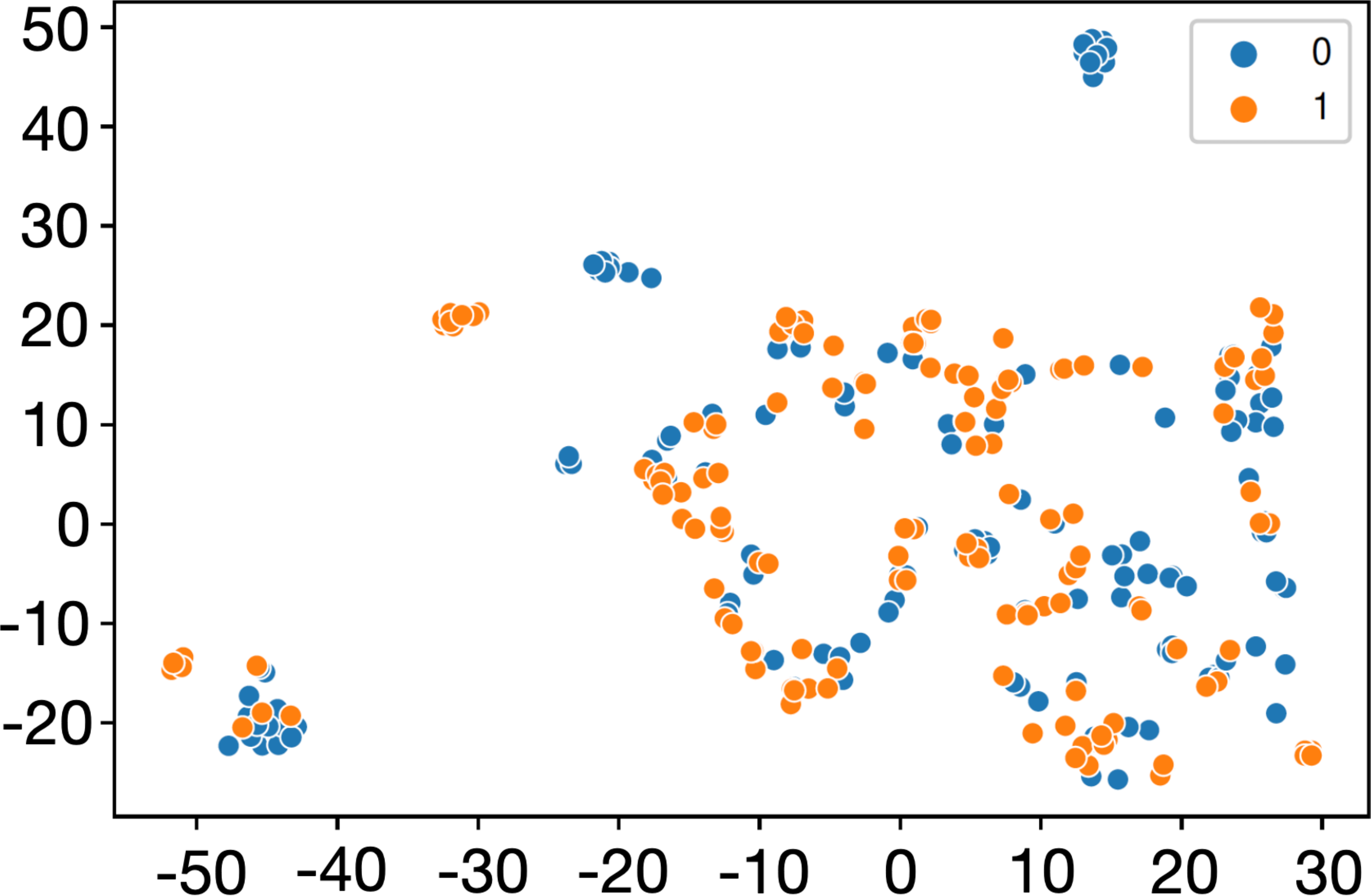}}\quad
\subfigure[GFKD]{\centering\label{fig:b}\includegraphics[height=2.6cm]{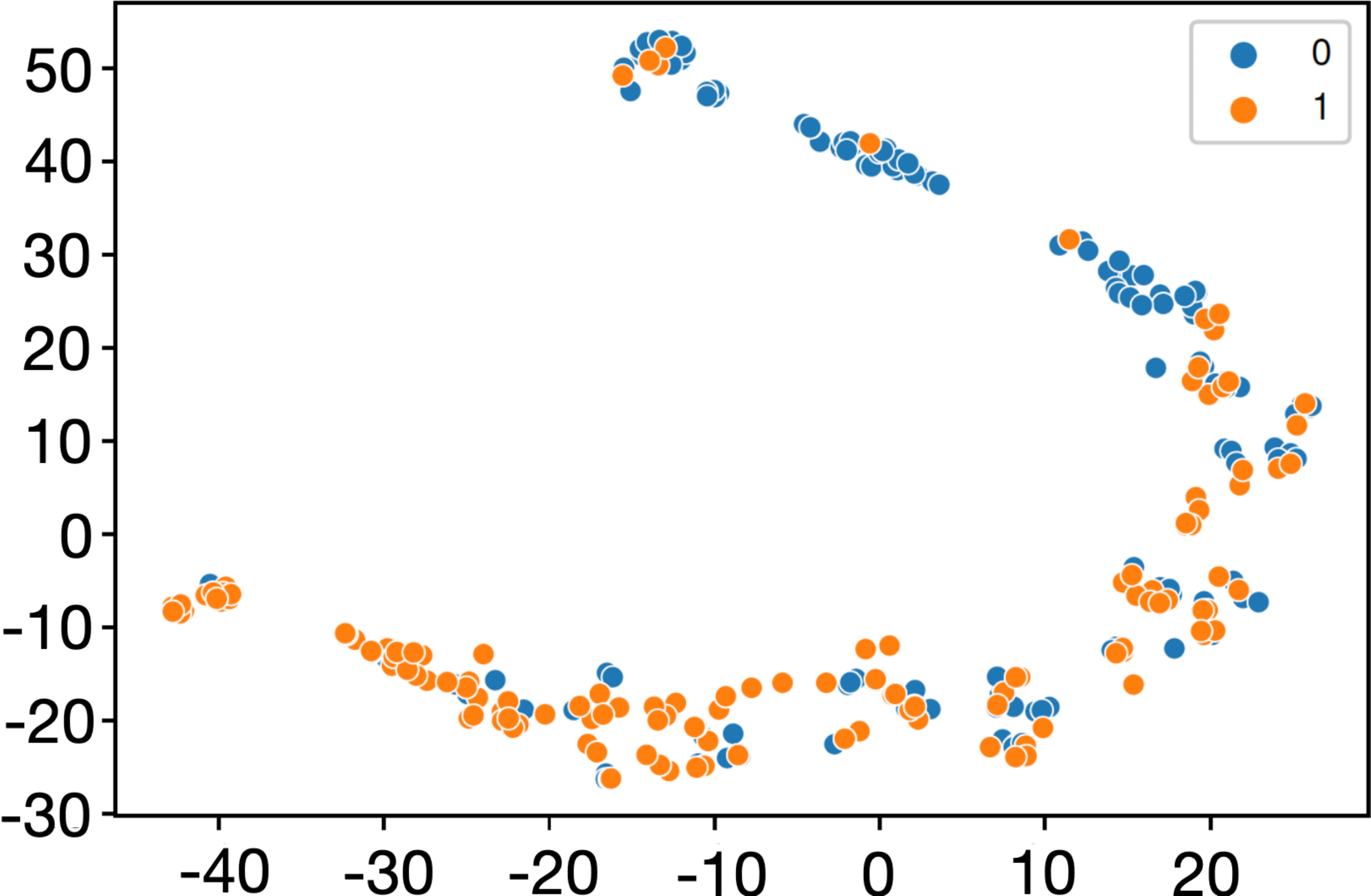}}\quad
\subfigure[Teacher]{\centering\label{fig:b}\includegraphics[height=2.6cm]{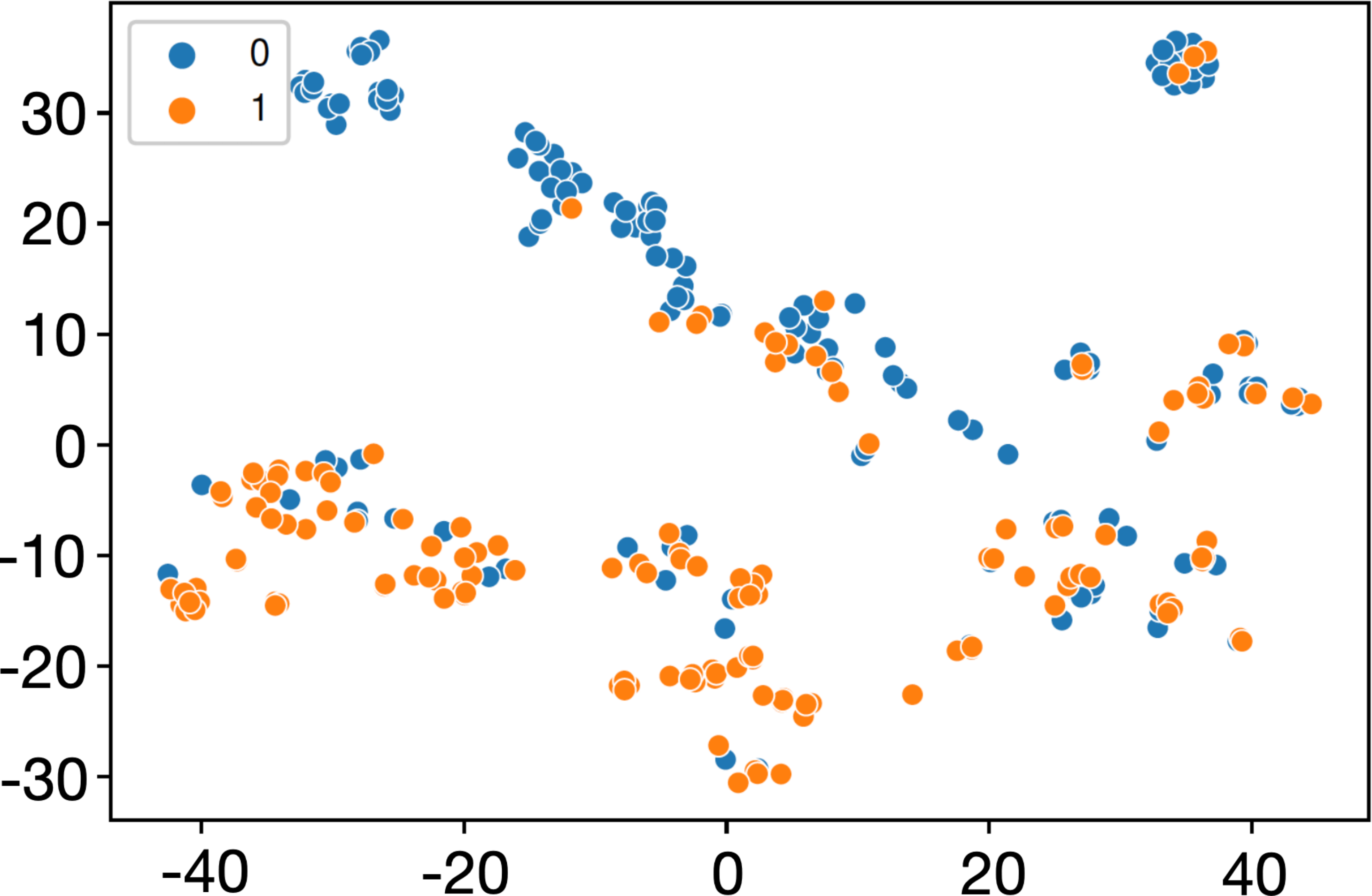}}
\vskip -0.1in
\caption{Feature visualization on IMDB-B}
\label{fea}
 \end{minipage}  
   \vskip -0.15in
\end{figure*}

\subsection{Experiments on Bioinformatics Graph Data}

Table \ref{bio} reports the comparison results on three bioinformatics graph datasets.
It is observed that without using any training data, GFKD transfers much more knowledge than those of the baselines across different datasets and architectures, which demonstrates the effectiveness of GFKD.
As expected, the overall performance of RandG is worse than those of the other methods as the knowledge in the teacher does not concentrate on random graphs.
GFKD and DeepInvG both learn node features for fake graphs, but differ in that GFKD learns graph structures while DeepInvG randomly generates structures.
As shown in Table \ref{bio}, GFKD improves the accuracy over DeepInvG substantially.
For example, the accuracy improvement of GFKD is 12.2\% over DeepInvG with teacher GCN-5-64 and student GCN-3-32 on MUTAG.
This demonstrates the effectiveness of GFKD for learning graph structures.
We also notice that in the teacher-student pair with different architectures, i.e, GCN-5-64 and GIN-3-32, GFKD also beats the baselines significantly on all the three datasets, which demonstrates that GFKD is applicable to the case where the teacher and the student have different architectures.

\subsection{Experiments on Social Network Graph Data}
To investigate the generalization of GFKD in different domains, we further conduct experiments on three social network graph datasets.
Table \ref{socail} reports the comparison results.
Note that on the three datasets, the node features are the degrees of the nodes (or a constant) which are derived from the graph structures.
Thus, DeepInvG is reduced to RandG.
It is observed that GFKD also outperforms the baselines substantially on all the three social network datasets, which demonstrates the generalization and usefulness of GFKD for different types of graph data.
The superiority of GFKD is attributed to its ability to learn the topology structures of graph data.

\subsection{Ablation Studies}


\subsubsection{Ablation Studies regarding the Number of Fake Graphs}
Theoretically, GFKD can generate infinite fake graphs for knowledge transfer.
However, the quality and the diversity are limited by the pretrained teacher.
We study how the performance of GFKD varies with the number of fake samples.
We denote the ratio of the number of fake graphs to the number of training samples used by the teacher by $r$.\par

Figure \ref{number} presents the effects of the number of fake graphs, where GCN-5-64 and GCN-3-32 are adopted as the teacher and the student, respectively.
It is not surprising that the performances of GFKD first increase and then stabilize. 
The reason for the performance stabilization is that the diversity of the generated graphs is constrained by the pretrained teacher.

\subsubsection{Ablation Studies regarding the Regularizers}
We have introduced two regularizers dealing with BN and one-hot features, respectively.
We evaluate their effects on the performances of GFKD by using GCN-5-64 and GCN-3-32 as the teacher and the student, respectively.
We adopt MUTAG and PTC datasets as their features are one-hot.
\par

The comparison results are presented in Figure \ref{aba}, where $\rm GFKD_{w/obn}$, $\rm GFKD_{w/ooh}$, and $\rm GFKD_{w/obnoh}$ denote GFKD without the BN regularizer, without the one-hot regularizer, and without neither of the two regularizers, respectively.
We observe that the performance decreases significantly without either of these two regularizers, which demonstrates the effectiveness and usefulness of these two regularizers.
Meanwhile, this also indicates that more prior knowledge about the graph data leads to better performances.

\subsection{Visualization}
Although the goal of GFKD is not to generate real graph data, we present some fake graphs learned by GFKD in Figure \ref{vis}, where the fake graphs are learned on IMDB-B from pretrained teacher GCN-5-64.
It is observed that the fake graphs and the real graphs share some visual similarities.
\par

To further investigate whether GFKD can learn discriminative features from these fake graphs, we use t-SNE~\cite{maaten2008visualizing} to visualize the features learned by different methods.
We adopt GCN-5-64 and GCN-3-32 as the teacher and the student in this experiment, respectively.
\par

Figure \ref{fea} presents the visualization of the features learned by RandG, GFKD, and the teacher.
It is observed that the feature representations learned by RandG are mixed for different classes, which indicates that using randomly generated graphs cannot learn discriminative features.
In contrast, the features learned by GFKD are well separated for different classes and are as discriminative as those learned by the teacher.
This demonstrates that the fake samples learned by GFKD are beneficial to representation learning and the knowledge concentrates on these fake graphs.

\section{Conclusion}

In this paper, we study a novel problem on how to distill knowledge from a GNN without observable graph data and introduce GFKD as a solution, which is to our best knowledge the first work along this line.
To learn where the knowledge in the teacher concentrates on, we propose to model the graph structures with multivariate Bernoulli distribution and then introduce a gradient estimator to optimize it.
Essentially, the structure gradients can be obtained by only using GNN forward propagation.
Extensive experiments on six benchmark datasets demonstrate the superiority of GFKD for extracting knowledge from GNNs without observable graphs.

\newpage
\bibliographystyle{named}
\bibliography{ijcai21}

\begin{thebibliography}{}

\bibitem[\protect\citeauthoryear{Bruna \bgroup \em et al.\egroup
  }{2013}]{bruna2013spectral}
Joan Bruna, Wojciech Zaremba, Arthur Szlam, and Yann LeCun.
\newblock Spectral networks and locally connected networks on graphs.
\newblock {\em arXiv preprint arXiv:1312.6203}, 2013.

\bibitem[\protect\citeauthoryear{Chen \bgroup \em et al.\egroup
  }{2019}]{chen2019data}
Hanting Chen, Yunhe Wang, Chang Xu, Zhaohui Yang, Chuanjian Liu, Boxin Shi,
  Chunjing Xu, Chao Xu, and Qi~Tian.
\newblock Data-free learning of student networks.
\newblock In {\em Proceedings of the IEEE International Conference on Computer
  Vision}, pages 3514--3522, 2019.

\bibitem[\protect\citeauthoryear{Defferrard \bgroup \em et al.\egroup
  }{2016}]{defferrard2016convolutional}
Micha{\"e}l Defferrard, Xavier Bresson, and Pierre Vandergheynst.
\newblock Convolutional neural networks on graphs with fast localized spectral
  filtering.
\newblock In {\em Advances in neural information processing systems}, pages
  3844--3852, 2016.

\bibitem[\protect\citeauthoryear{Fey \bgroup \em et al.\egroup
  }{2018}]{fey2018splinecnn}
Matthias Fey, Jan Eric~Lenssen, Frank Weichert, and Heinrich M{\"u}ller.
\newblock Splinecnn: Fast geometric deep learning with continuous b-spline
  kernels.
\newblock In {\em Proceedings of the IEEE Conference on Computer Vision and
  Pattern Recognition}, pages 869--877, 2018.

\bibitem[\protect\citeauthoryear{Goodfellow \bgroup \em et al.\egroup
  }{2014}]{goodfellow2014generative}
Ian Goodfellow, Jean Pouget-Abadie, Mehdi Mirza, Bing Xu, David Warde-Farley,
  Sherjil Ozair, Aaron Courville, and Yoshua Bengio.
\newblock Generative adversarial nets.
\newblock In {\em Advances in neural information processing systems}, pages
  2672--2680, 2014.

\bibitem[\protect\citeauthoryear{Hamilton \bgroup \em et al.\egroup
  }{2017}]{hamilton2017inductive}
Will Hamilton, Zhitao Ying, and Jure Leskovec.
\newblock Inductive representation learning on large graphs.
\newblock In {\em Advances in neural information processing systems}, pages
  1024--1034, 2017.

\bibitem[\protect\citeauthoryear{Hinton \bgroup \em et al.\egroup
  }{2015}]{hinton2015distilling}
Geoffrey Hinton, Oriol Vinyals, and Jeff Dean.
\newblock Distilling the knowledge in a neural network.
\newblock {\em arXiv preprint arXiv:1503.02531}, 2015.

\bibitem[\protect\citeauthoryear{Ioffe and Szegedy}{2015}]{ioffe2015batch}
Sergey Ioffe and Christian Szegedy.
\newblock Batch normalization: Accelerating deep network training by reducing
  internal covariate shift.
\newblock {\em arXiv preprint arXiv:1502.03167}, 2015.

\bibitem[\protect\citeauthoryear{Kipf and Welling}{2017}]{kipf2016semi}
Thomas~N Kipf and Max Welling.
\newblock Semi-supervised classification with graph convolutional networks.
\newblock {\em International Conference on Learning Representations}, 2017.

\bibitem[\protect\citeauthoryear{Li \bgroup \em et al.\egroup
  }{2015}]{li2015gated}
Yujia Li, Daniel Tarlow, Marc Brockschmidt, and Richard Zemel.
\newblock Gated graph sequence neural networks.
\newblock {\em arXiv preprint arXiv:1511.05493}, 2015.

\bibitem[\protect\citeauthoryear{Lopes \bgroup \em et al.\egroup
  }{2017}]{lopes2017data}
Raphael~Gontijo Lopes, Stefano Fenu, and Thad Starner.
\newblock Data-free knowledge distillation for deep neural networks.
\newblock {\em arXiv preprint arXiv:1710.07535}, 2017.

\bibitem[\protect\citeauthoryear{Maaten and
  Hinton}{2008}]{maaten2008visualizing}
Laurens van~der Maaten and Geoffrey Hinton.
\newblock Visualizing data using t-sne.
\newblock {\em Journal of machine learning research}, 9(Nov):2579--2605, 2008.

\bibitem[\protect\citeauthoryear{Micaelli and Storkey}{2019}]{micaelli2019zero}
Paul Micaelli and Amos~J Storkey.
\newblock Zero-shot knowledge transfer via adversarial belief matching.
\newblock In {\em Advances in Neural Information Processing Systems}, pages
  9551--9561, 2019.

\bibitem[\protect\citeauthoryear{Mordvintsev \bgroup \em et al.\egroup
  }{}]{mordvintsev2015inceptionism}
Alexander Mordvintsev, Christopher Olah, and Mike Tyka.
\newblock Inceptionism: Going deeper into neural networks.
\newblock https://ai.googleblog.com/2015/06/
  inceptionism-going-deeper-into-neural.html, last accessed on 12.20.2020.

\bibitem[\protect\citeauthoryear{Nayak \bgroup \em et al.\egroup
  }{2019}]{nayak2019zero}
Gaurav~Kumar Nayak, Konda~Reddy Mopuri, Vaisakh Shaj, R~Venkatesh Babu, and
  Anirban Chakraborty.
\newblock Zero-shot knowledge distillation in deep networks.
\newblock {\em arXiv preprint arXiv:1905.08114}, 2019.

\bibitem[\protect\citeauthoryear{Romero \bgroup \em et al.\egroup
  }{2015}]{romero2014fitnets}
Adriana Romero, Nicolas Ballas, Samira~Ebrahimi Kahou, Antoine Chassang, Carlo
  Gatta, and Yoshua Bengio.
\newblock Fitnets: Hints for thin deep nets.
\newblock In {\em International Conference on Learning Representations}, 2015.

\bibitem[\protect\citeauthoryear{Tian \bgroup \em et al.\egroup
  }{2020}]{tian2019contrastive}
Yonglong Tian, Dilip Krishnan, and Phillip Isola.
\newblock Contrastive representation distillation.
\newblock In {\em International Conference on Learning Representations}, 2020.

\bibitem[\protect\citeauthoryear{Veli{\v{c}}kovi{\'c} \bgroup \em et al.\egroup
  }{2017}]{velivckovic2017graph}
Petar Veli{\v{c}}kovi{\'c}, Guillem Cucurull, Arantxa Casanova, Adriana Romero,
  Pietro Lio, and Yoshua Bengio.
\newblock Graph attention networks.
\newblock {\em arXiv preprint arXiv:1710.10903}, 2017.

\bibitem[\protect\citeauthoryear{Williams}{1992}]{williams1992simple}
Ronald~J Williams.
\newblock Simple statistical gradient-following algorithms for connectionist
  reinforcement learning.
\newblock {\em Machine learning}, 8(3-4):229--256, 1992.

\bibitem[\protect\citeauthoryear{Xu \bgroup \em et al.\egroup
  }{2019}]{xu2018powerful}
Keyulu Xu, Weihua Hu, Jure Leskovec, and Stefanie Jegelka.
\newblock How powerful are graph neural networks?
\newblock {\em International Conference on Learning Representations}, 2019.

\bibitem[\protect\citeauthoryear{Yang \bgroup \em et al.\egroup
  }{2020}]{yang2020distilling}
Yiding Yang, Jiayan Qiu, Mingli Song, Dacheng Tao, and Xinchao Wang.
\newblock Distilling knowledge from graph convolutional networks.
\newblock In {\em Proceedings of the IEEE/CVF Conference on Computer Vision and
  Pattern Recognition}, pages 7074--7083, 2020.

\bibitem[\protect\citeauthoryear{Yin and Zhou}{2019}]{yin2019arm}
Mingzhang Yin and Mingyuan Zhou.
\newblock Arm: Augment-reinforce-merge gradient for stochastic binary networks.
\newblock In {\em International Conference on Learning Representations}, 2019.

\bibitem[\protect\citeauthoryear{Yin \bgroup \em et al.\egroup
  }{2020}]{yin2020dreaming}
Hongxu Yin, Pavlo Molchanov, Jose~M Alvarez, Zhizhong Li, Arun Mallya, Derek
  Hoiem, Niraj~K Jha, and Jan Kautz.
\newblock Dreaming to distill: Data-free knowledge transfer via deepinversion.
\newblock In {\em Proceedings of the IEEE/CVF Conference on Computer Vision and
  Pattern Recognition}, pages 8715--8724, 2020.

\bibitem[\protect\citeauthoryear{Yoo \bgroup \em et al.\egroup
  }{2019}]{yoo2019knowledge}
Jaemin Yoo, Minyong Cho, Taebum Kim, and U~Kang.
\newblock Knowledge extraction with no observable data.
\newblock In {\em Advances in Neural Information Processing Systems}, pages
  2705--2714, 2019.

\bibitem[\protect\citeauthoryear{Zagoruyko and
  Komodakis}{2017}]{zagoruyko2016paying}
Sergey Zagoruyko and Nikos Komodakis.
\newblock Paying more attention to attention: Improving the performance of
  convolutional neural networks via attention transfer.
\newblock In {\em International Conference on Learning Representations}, 2017.

\end{thebibliography}

\end{document}